\documentclass[10pt,twocolumn,letterpaper]{article}

\usepackage{iccv}
\usepackage{times}
\usepackage{epsfig}
\usepackage{graphicx}
\usepackage{amsmath}
\usepackage{amssymb}
\usepackage{multirow}
\usepackage{subfigure}

\usepackage[breaklinks=true,bookmarks=false]{hyperref}

\iccvfinalcopy 

\def\iccvPaperID{5215} 

\ificcvfinal\fi

\begin{document}

\setcounter{footnote}{-1}
\title{Two-Stream Action Recognition-Oriented Video Super-Resolution
\thanks{This work was supported by the National Key Research and Development Plan under Grant 2017YFB1002401, and by the Natural Science Foundation of China under Grant 61772483. \emph{(Corresponding author: Dong Liu.)}}
}

\author{Haochen Zhang, Dong Liu, Zhiwei Xiong\\
CAS Key Laboratory of Technology in Geo-Spatial Information Processing and Application System,\\
University of Science and Technology of China, Hefei 230027, China\\
{\tt\small zhc12345@mail.ustc.edu.cn, \{dongeliu, zwxiong\}@ustc.edu.cn}
}

\maketitle
\ificcvfinal\thispagestyle{empty}\fi

\begin{abstract}
We study the video super-resolution (SR) problem for facilitating video analytics tasks, \eg action recognition, instead of for visual quality. The popular action recognition methods based on convolutional networks, exemplified by two-stream networks, are not directly applicable on video of low spatial resolution. This can be remedied by performing video SR prior to recognition, which motivates us to improve the SR procedure for recognition accuracy. Tailored for two-stream action recognition networks, we propose two video SR methods for the spatial and temporal streams respectively. On the one hand, we observe that regions with action are more important to recognition, and we propose an optical-flow guided weighted mean-squared-error loss for our spatial-oriented SR (SoSR) network to emphasize the reconstruction of moving objects. On the other hand, we observe that existing video SR methods incur temporal discontinuity between frames, which also worsens the recognition accuracy, and we propose a siamese network for our temporal-oriented SR (ToSR) training that emphasizes the temporal continuity between consecutive frames. We perform experiments using two state-of-the-art action recognition networks and two well-known datasets--UCF101 and HMDB51. Results demonstrate the effectiveness of our proposed SoSR and ToSR in improving recognition accuracy.
\end{abstract}

\section{Introduction}

In recent years, convolutional neural networks (CNNs) have been applied to action recognition task and obtained state-of-the-art performance over the traditional arts. For the convenience of classification, most of them adopt fully-connected layers in their architecture and thus these well-trained CNNs cannot be directly applied on low-resolution (LR) video. As a result, their widely application could be hindered by video's resolution. Most of the datasets used for studying action recognition have a fixed resolution, \eg UCF101 (about 320$\times$240), HMDB51 (about 340$\times$256), Sports-1M (about 640$\times$360) \cite{herath2017going}. But the resolution in real world usually varies among different sources of video capturing, inevitably being low, \eg in surveillance scenario. There are also many situations where the video has high resolution but the region containing action is quite small. 

To address the resolution problem, most of existing works choose interpolation to simply re-scale the input while some recent works in other areas propose super-resolution (SR) as an alternative solution. For example, \cite{shermeyer2018effects} investigated the effects of SR on object detection and \cite{vidal2018ug} proposed a dataset for assessing the impact of image restoration and enhancement on image classification. Most of them stop at a preliminary experimental study on existing SR methods without proposing approaches targeting on their tasks, \eg action recognition.

Super-resolution, aiming to enhance the resolution of images or video, has long attracted the attention of researchers. In early years, the target of enhancement is mainly signal fidelity, \eg PSNR, partly because of an intuitive assumption that PSNR is consistent with visual quality, and mean-squared-error (MSE) is extensively used during optimization. However, some recent works challenge this assumption by showing distortion and perception can be tradeoff \cite{blau2018perception}, and introduce some perceptual loss \cite{johnson2016perceptual,ledig2017photo} in addition to MSE, targeting for better visual quality of super-resolved image. Nonetheless, it is still not clear whether visual quality determines the quality of visual analytics results, \eg action recognition accuracy. Since the analytics tasks are performed by computer instead of human, we argue that the SR methods optimized for visual quality may not be optimal for action recognition task.

We investigate the video SR problem aiming to facilitate recognition quality, exemplified by action recognition, instead of visual quality. In particular, we use SR as a preprocessing step before feeding LR video into an action recognition network that is well-trained on HR video. We investigate recognition quality of different super-resolved video evaluated by a computer algorithm rather than by human. The problem we want to address is that given a data-driven classifier, exemplified by an action recognition CNN, with parameters trained on HR video, what the accuracy will be when the same classifier deals with LR video assisted with SR preprocessing.  In addition, we want to find how to design a better SR network targeting on recognition to improve the accuracy.

Oriented to the popular two-stream action recognition framework \cite{simonyan2014two} which learns two separate networks, one for spatial color information and the other for temporal motion information, we propose two video SR methods for these two streams respectively. For the spatial stream which can be regarded as image classification, we observe that the moving object is more related to the recognition and should be paid more attention during SR enhancement. Thus, our Spatial-oriented SR (SoSR) takes weighted mean-squared-error guided by optical flow as loss to emphasize moving objects. For the temporal stream, we observe that video SR can result in the temporal discontinuity between consecutive video frames which may harm the quality of optical flow and incur drop in recognition accuracy. Thus, in our Temporal-oriented SR (ToSR), we enhance the consecutive frames together to ensure the temporal consistency.

Our contributions can be summarized as follows. 
We investigate state-of-the-art image and video SR methods from the view of facilitating action recognition, assuming well-trained two-stream networks as ``evaluators.''
For the spatial stream, we propose an optical flow guided weighted MSE loss to guide our SoSR to pay more attention to regions with motion. 
For the temporal stream, we propose ToSR which enhances the consecutive frames together to achieve temporal consistency.

We perform experiments with two state-of-the-art recognition networks on two widely used datasets--UCF101 \cite{soomro2012ucf101} and HMDB51 \cite{kuehne2011hmdb}. Comprehensive experimental results show that our SoSR and ToSR indeed improve the recognition accuracy significantly. Especially on the HMDB51 dataset, our proposed method can improve the recognition performance of LR video from 42.81\% to 53.59\% on spatial stream and from 56.54\% to 61.5\% on temporal stream. Our code is released\footnote{\url{https://github.com/AlanZhang1995/TwoStreamSR}}.
\section{Related Work}
We review related works at two aspects: action recognition and image/video SR. In both fields, CNN has been the mainstream and outperforms the traditional methods significantly. Thus we only mention several CNN-based approaches that are highly related to our work.

\begin{figure*}
	\begin{center}
		\includegraphics[width=\linewidth]{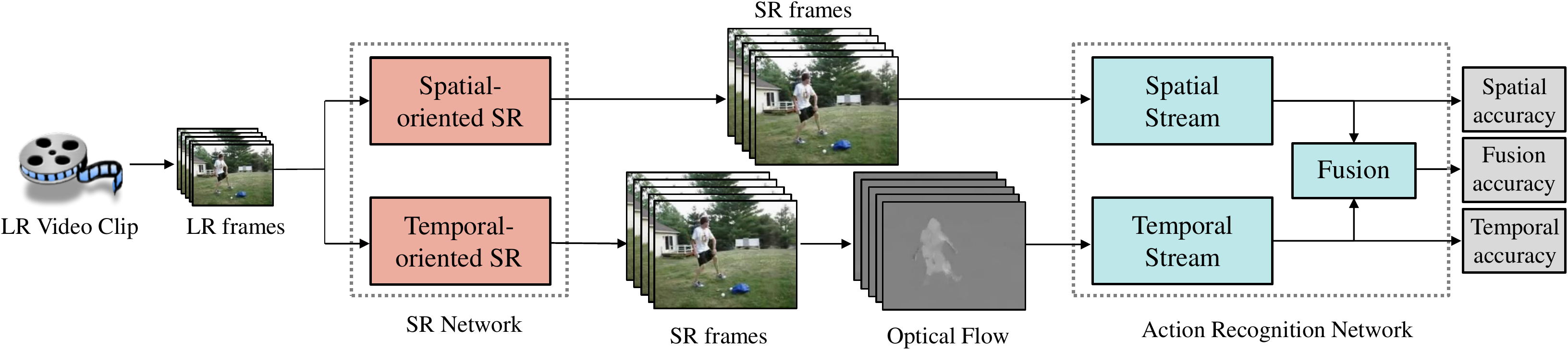}
	\end{center}
	\caption{The pipeline of performing video SR prior to action recognition for LR video. Note that our work focuses on the SR network, we directly adopt the well-trained two-stream action recognition network without any tuning.}
	\label{fig:pipeline}
	\vspace{-5pt}
\end{figure*}

\textbf{CNN for action recognition.} In CNN-based action recognition, a key problem is how to properly incorporate spatial and temporal information in CNN architectures. Solutions can be divided into three categories: 3D convolution, RNN/LSTM, and two-stream. 3D CNN which learns spatio-temporal features was first presented in \cite{ji20133d}. Later on, C3D features and 3D CNN architectures \cite{diba2017deep,tran2015learning,varol2018long,wang2018two} appeared. There were also several works \cite{qiu2017learning,sun2015human,zhou2018mict} focusing on improvements of 3D CNNs. RNN/LSTM is believed to cope with sequential information better, and thus \cite{donahue2015long,wu2015modeling,yue2015beyond} attempted to incorporate LSTMs to deal with action recognition. Two-stream CNN architecture was firstly proposed in \cite{simonyan2014two}. This architecture consists of two separate networks, one for exploiting spatial information from individual frames, and the other for using temporal information from optical flow; the outputs of two networks are then combined by late fusion. Several improvements were presented for two-stream \cite{feichtenhofer2016spatiotemporal,feichtenhofer2016convolutional,wang2016temporal}. We design SR methods specifically for two-stream networks due to two reasons. First, two-stream approach seems leading to the best performance for action recognition on several benchmarks. Second, both 3D convolution and RNN/LSTM networks are not easily decomposed, but two-stream networks have a clear decomposition, which facilitates the investigation of SR. We use two state-of-the-art methods, Temporal Segment Network (TSN) \cite{wang2016temporal} and Spatio-Temporal Residual Network (ST-Resnet) \cite{feichtenhofer2016spatiotemporal}, in our experiments.

\textbf{CNN for image SR.} Almost all of the existing image SR methods are designed to enhance the visual quality by adding more image details. In earlier years, PSNR is evaluated as a surrogate of visual quality and thus mean-squared-error is extensively used as loss function \cite{dong2016image,kim2016accurate,kim2016deeply,lim2017enhanced,liu2019single,shi2016real,tai2017image,tai2017memnet,zhang2018image}. More recently, visual quality is considered directly and several different kinds of loss functions are proposed, such as feature loss \cite{johnson2016perceptual} and loss defined by generative adversarial network (GAN) \cite{goodfellow2014generative}. For example, Ledig \etal \cite{ledig2017photo} proposed SRGAN which combined GAN loss and feature loss. It is also worth noting that PSNR and visual quality can be even contradictory \cite{blau2018perception}.

\textbf{CNN for video SR.} Compared to single image SR, the temporal dimension provides much more information in video SR, and various methods have been proposed to exploit the temporal information \cite{liu2017robust,liu2018learning}.
A majority of these methods have an explicit motion compensation module to align different frames. For example, Kappeler \etal \cite{kappeler2016video} slightly modified SRCNN \cite{dong2016image} and extracted features from frames that were aligned by optical flow. Caballero \etal \cite{caballero2017real} proposed an end-to-end SR network to learn motions between input LR frames and generate SR frames in real time. Tao \etal \cite{tao2017detail} introduced a new sub-pixel motion compensation (SPMC) layer to perform motion compensation and up-sampling jointly.
Also several methods try to avoid the explicit motion compensation. For example, Jo \etal \cite{jo2018deep} proposed a network that used dynamic up-sampling filters.
All the aforementioned works are pursuing higher PSNR for video SR. But we consider video SR to improve action recognition accuracy. We focus on the loss functions instead of the network structures.
\section{Action Recognition-Oriented SR}

\textbf{Pipeline.} Figure \ref{fig:pipeline} depicts the pipeline of using SR for action recognition by two-stream networks. Given an LR video sequence, we split it into frames on which we perform SR enhancement. We propose Spatial-oriented SR (SoSR) and Temporal-oriented SR (ToSR) for the two streams, \ie we enhance the LR video twice. We then calculate optical flow from the ToSR resulting video, and feed the optical flow together with frames from the SoSR resulting video into the following recognition network.

\textbf{Action Recognition Network.} Our SR methods are specifically designed for two-stream action recognition networks. Specifically, we use TSN \cite{wang2016temporal} and ST-Resnet \cite{feichtenhofer2016spatiotemporal} in our experiments. There are minor differences between the two networks: TSN uses a weighted average of the classification scores predicted from the two streams, while ST-Resnet trains a fusion sub-network together with the two streams in an end-to-end fashion. We focus on the SR part and directly use the well-trained models provided by the authors \textit{without any tuning}.

\textbf{End-to-End Optimization?} According to Figure \ref{fig:pipeline}, it appears an appealing choice to perform an end-to-end optimization, i.e. training the SR network with the recognition accuracy as the objective. However, this involves a specific action recognition model. Our empirical results indicate that, training the SR with a specific action recognition model (e.g. TSN), and testing with another model (e.g. ST-Resnet), leads to much worse results. It motivates us to design specific loss functions for the SR training.
\subsection{Spatial-oriented SR}
\subsubsection{Analysis}
\begin{figure}
	\centering
	\subfigure[10-th frame of Archery\_g01\_c07]{
		\begin{minipage}[c]{\linewidth}
			\includegraphics[width=\columnwidth]{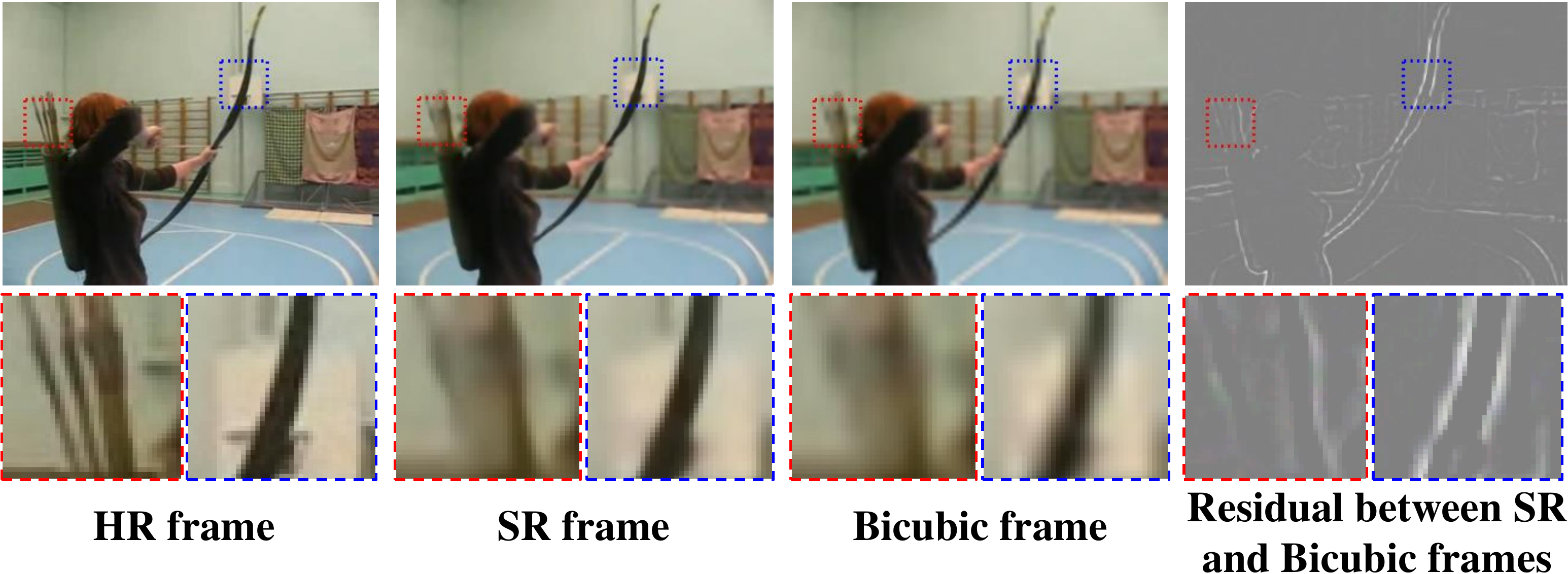}
	\end{minipage}}
	\subfigure[152-nd frame of JumpRope\_g02\_c02]{
		\begin{minipage}[c]{\linewidth}
			\includegraphics[width=\columnwidth]{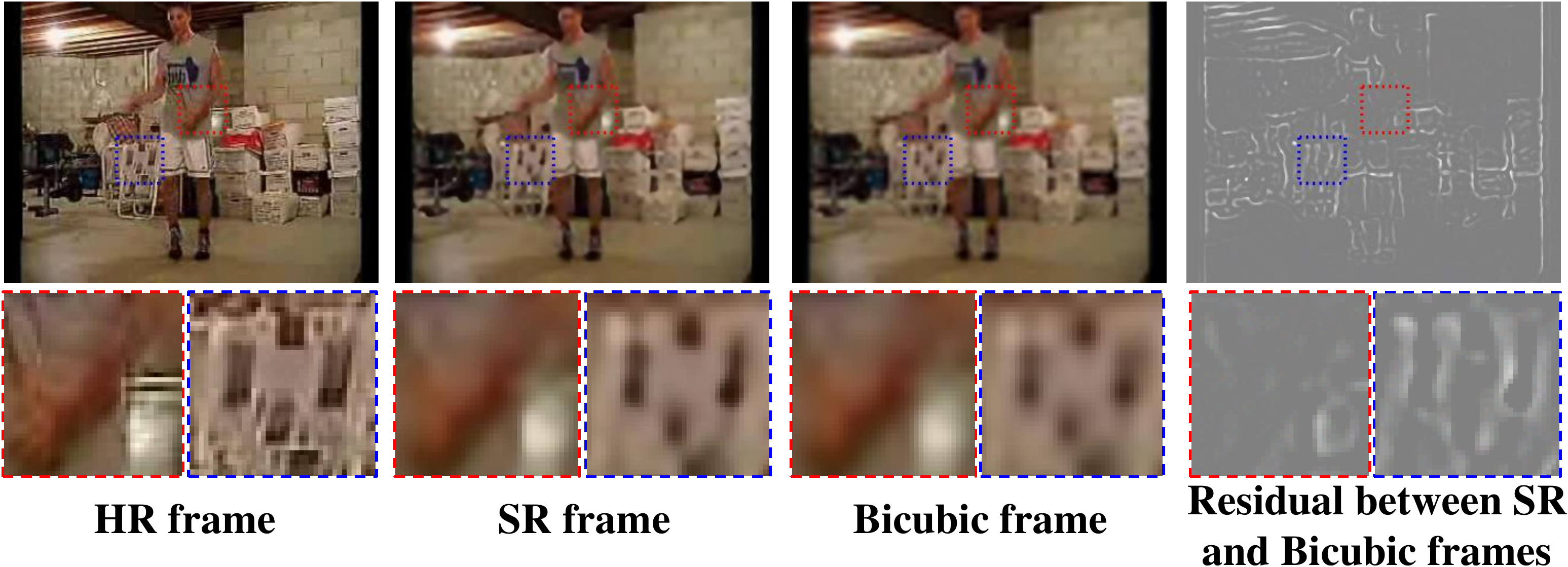}
	\end{minipage}}
	\subfigure[39-th frame of HandstandWalking\_g06\_c01]{
		\begin{minipage}[c]{\linewidth}
			\includegraphics[width=\columnwidth]{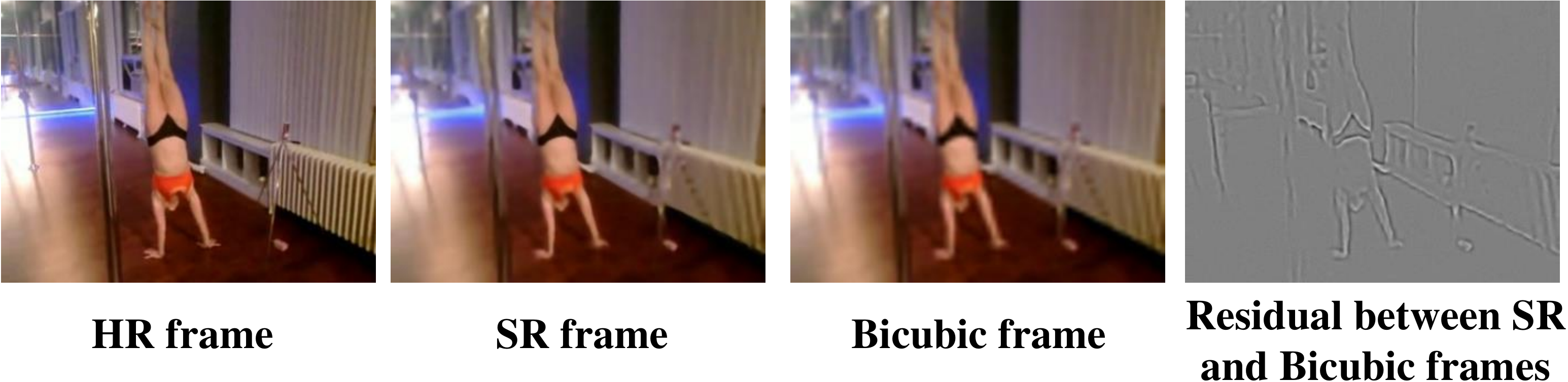}
	\end{minipage}}
	\caption{Examples show how image details added by VDSR \cite{kim2016accurate} influence the recognition accuracy. In (a), VDSR adds details on the bow, and since bow directly relates to \texttt{Archery}, VDSR improves recognition than bicubic. In (b), VDSR adds details on the background but not on the object, resulting in even lower accuracy than bicubic. In (c), VDSR adds details on the walking woman but not on the background, resulting in even higher accuracy than HR. See Table \ref{tab:SoSRAnalysis} for the accuracy values.}
	\label{fig:insight}
\end{figure}

According to the two-stream architecture, the spatial stream performs recognition from individual frames by recognizing objects. That says, the spatial stream is equivalent to image classification. Inspired by previous work \cite{dai2016image}, we expect that SR can enhance the LR frames and add more image details with which SR helps in recognition. However, when we experiment with a representative image SR method, namely VDSR \cite{kim2016accurate}, we observe some counterexamples. We calculate recognition accuracy for individual classes, and find that VDSR sometimes performs worse than the simple bicubic interpolation, more interestingly, the original HR frames can be worse than super-resolved or even interpolated frames. Such examples are summarized in Table \ref{tab:SoSRAnalysis} and Figure \ref{fig:insight}. 
Considering that LR frames lose details compared to HR frames, bicubic interpolation simply up-scales frames without adding details, and SR methods usually enhance interpolated frames with image details, we conjecture that, especially in specific classes, image details can be either helpful or harmful for action recognition depending on the regions where details are added.

In Figure \ref{fig:insight}, we visually analyze some frames to confirm our conjecture. In (a), which corresponds to HR$>$VDSR$>$Bicubic, we indeed observe that many details about the bow and arrow lie in the HR frame, but are missing in bicubic frame; the SR frame adds some details on the bow (shown in the blue box), which is helpful for recognition since the bow is directly related to the class \texttt{Archery}. In (b), which corresponds to Bicubic$>$HR$>$VDSR, we observe that SR frame contains more details than bicubic frame, mostly on the background (shown in the blue box) rather than on the key object (shown in the red box); the added details seem to be harmful for action recognition. In (c), which corresponds to VDSR$>$HR$>$Bicubic, as the SR frame has more details on the human (object directly related to \texttt{Walking}) but fewer details on the background (due to LR input), the recognition accuracy is even higher than HR.

\begin{table}
	\begin{center}
	\footnotesize\selectfont
			\begin{tabular}{|c|c|c|c|c|}
				\hline
				\multirow{2}*{Case} & \multirow{2}*{Class} &  \multicolumn{3}{c|}{Recognition Accuracy (\%)} \\
				\cline{3-5}
				& & HR & Bicubic & VDSR \\
				\hline
				\multirow{2}*{a} & Archery & \textbf{82.93} & 36.59 & 70.73 \\
				\cline{2-5}
				& PlayingFlute & \textbf{97.92} & 72.92 & 79.17 \\
				\hline
				\multirow{2}*{b} & JumpRope & 39.47 & \textbf{42.11} & 7.89 \\
				\cline{2-5}
				& SalsaSpin & 79.07 & \textbf{83.72} & 53.49 \\
				\hline
				\multirow{2}*{c} & FrontCrawl & 64.86 & 32.43 & \textbf{78.38} \\
				\cline{2-5}
				& HandstandWalking & 35.29 & 29.41 & \textbf{41.18} \\
				\hline
		\end{tabular}
	\end{center}
	\caption{We observe different cases in recognition accuracy for the classes in UCF101 using the TSN network. In case (a), HR$>$VDSR$>$Bicubic. In case (b), Bicubic$>$HR$>$VDSR. In case (c), VDSR$>$HR$>$Bicubic. Some representative classes are presented in this table. See Figure \ref{fig:insight} for visual inspection. (The scaling factor is 4 throughout the entire paper unless otherwise indicated.)}
	\label{tab:SoSRAnalysis}
\end{table}

It is worth noting that, despite of these counterexamples, HR video is still the best in terms of recognition accuracy on the overall sense (as shown in Tables \ref{tab:Table1}), partly because the action recognition networks are trained on HR frames. What if the networks are trained with LR video? We will study in Section \ref{subsec_visual}.

\subsubsection{Method}
Based on the observation, we propose an SR method to selectively enhance the image regions that are highly related to action recognition. These regions usually have high motion, such as the bow in Figure \ref{fig:insight} (a), the rotating rope in Figure \ref{fig:insight} (b), and the walking woman in Figure \ref{fig:insight} (c). We select these regions according to the optical flow since optical flow is a commonly chosen representation for motion information. Note that high motion does not necessarily relate to action. It is a much simplified implementation, but seems working well in our experiments.

Most of SR networks use mean-squared-error (MSE) as loss function, which is to assume equal importance of every pixel. In contrast, we propose to use a weighted MSE (WMSE) based on optical flow to emphasize some pixels that are more important than others. In short, the loss function we used here is
\begin{equation}
\label{wmse1}
\textrm{WMSE}=\frac{1}{N}{\sum_{p=1}^{N}\left \| I(p)-\hat{I}(p) \right \|^{2}\cdot \sqrt{u^{2}(p)+v^{2}(p)}},
\end{equation}
where $I$ and $\hat{I}$ are HR and SR frames respectively, $p$ is the pixel index, and $N$ is the total number of pixels. $u$ and $v$ represent the magnitude of optical flow in the horizontal and vertical directions respectively. Here, the optical flow is calculated offline from the HR video using Flownet 2.0 \cite{ilg2017flownet}, which we observe is slightly better than using TVL$^1$ \cite{zach2007duality}. In this way, the loss can guide the network in a pixel-wise manner: pixels with larger motion correspond to larger loss weights and thus are paid more attention during SR enhancement.

\begin{table}
	\begin{center}
	\footnotesize\selectfont
			\begin{tabular}{|c|c|c|c|c|}
				\hline  
				Structure &	MSE/WMSE &	Feature & Adversarial & Accuracy \\
				\hline
				VDSR & MSE  &  - & -  & 46.6\% \\  
				\hline  
				VDSR & WMSE & -  &  - & 47.91\% \\ 
				\hline 
				VDSR & WMSE &  \checkmark &  -  & 50.39\% \\  
				\hline 
				ESRGAN & WMSE &  \checkmark &  - &  52.55\% \\  
				\hline  
				ESRGAN & MSE &  \checkmark & \checkmark & 52.48\% \\ 
				\hline 
				ESRGAN & WMSE &  \checkmark & \checkmark  & \textbf{53.59\%} \\  
				\hline  
		\end{tabular}
	\end{center}
	\caption{Ablation study for SoSR using different network structures and different loss functions, with TSN \cite{wang2016temporal} on HMDB51 dataset.}
	\label{tab:SoSR_Ablation}
\end{table}

In addition to WMSE, we further investigate two kinds of perceptual loss: feature loss and adversarial loss, which have been widely used in recent SR methods for improving visual quality \cite{johnson2016perceptual,ledig2017photo}. Using feature loss is to minimize the difference of high-level image features between SR image and HR image and using adversarial loss is to generate SR image which is closer to HR image in terms of distribution. 

As mentioned before, the spatial stream is equivalent to image classification. We anticipate that single frame SR can perform well for the spatial stream and also has lower complexity than multi-frame SR. So here, we investigate two image SR network structures: One is VDSR \cite{kim2016accurate} structure and the other is based on ESRGAN \cite{wang2018esrgan}.

We conduct an ablation study about the proposed loss function and different network structures. As shown in Table \ref{tab:SoSR_Ablation}, feature loss, adversarial loss as well as advanced network structure are all beneficial to the final recognition accuracy, and our proposed WMSE further improves the recognition performance. For more analyses please refer to Section \ref{subsec_res}.

As a result, we train our SoSR network based on ESRGAN with our training data (details in Section \ref{subsec_exp}) and the following loss function:
\begin{equation}
\label{SoSR_loss}
\mathcal{L}_{\textrm{SoSR}}= \alpha \mathcal{L}_{\textrm{WMSE}}+\beta \mathcal{L}_{\textrm{Feature}}+\gamma \mathcal{L}_{\textrm{Adversarial}},
\end{equation}
where $\alpha$, $\beta$ and $\gamma$ are weights.

\subsection{Temporal-oriented SR}
\subsubsection{Analysis}
\begin{figure}[t]
	\begin{center}
		\includegraphics[width=\linewidth]{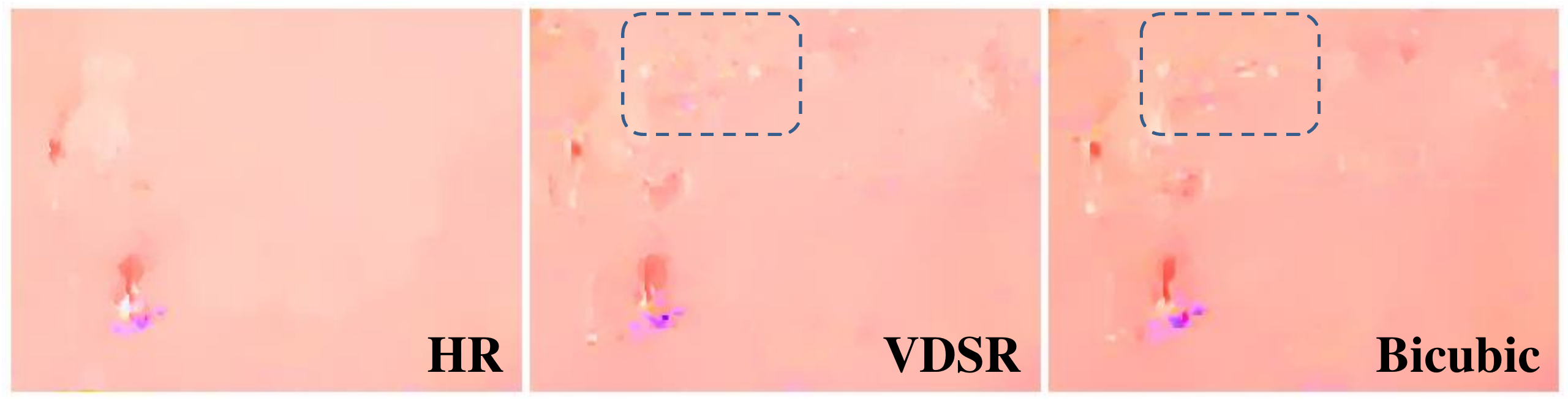}
	\end{center}
	\caption{Optical flow maps calculated from HR, VDSR \cite{kim2016accurate}, and bicubic video, respectively. Artifacts can be found in the circled regions. Zooming-in inspection can observe that SR video has more artifacts than bicubic one. In this example, VDSR incurs lower recognition accuracy than bicubic on the temporal stream.}
	\label{fig:opticalflow}
\end{figure}

We now switch to the temporal stream. As described in the two-stream architecture, the temporal stream takes optical flow as input to utilize temporal information. 
Thus, the question should be how SR affects the quality of optical flow. We again experiment with the representative image SR method--VDSR \cite{kim2016accurate}. Figure \ref{fig:opticalflow} shows the optical flow maps calculated from HR video, SR video, and bicubic interpolated video, respectively. Here the optical flow is calculated by the TVL$^1$ method \cite{zach2007duality} as most action recognition works do.
From Figure \ref{fig:opticalflow}, we can find that the optical flow from bicubic video has a lot of artifacts, and VDSR even worsens the optical flow. Thus, VDSR incurs less appealing results of recognition accuracy.

\begin{figure}
	\centering
	\subfigure[A video sequence (TaiChi\_g01\_c04)]{
		\begin{minipage}[c]{\linewidth}
			\includegraphics[width=\columnwidth]{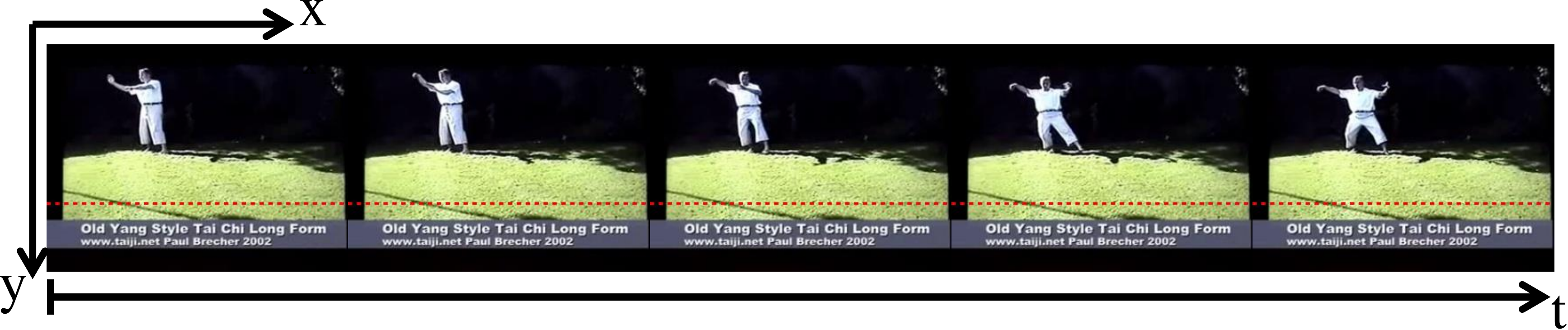}
	\end{minipage}}
	\subfigure[Temporal profiles of different results]{
		\begin{minipage}[c]{\linewidth}
			\includegraphics[width=\columnwidth]{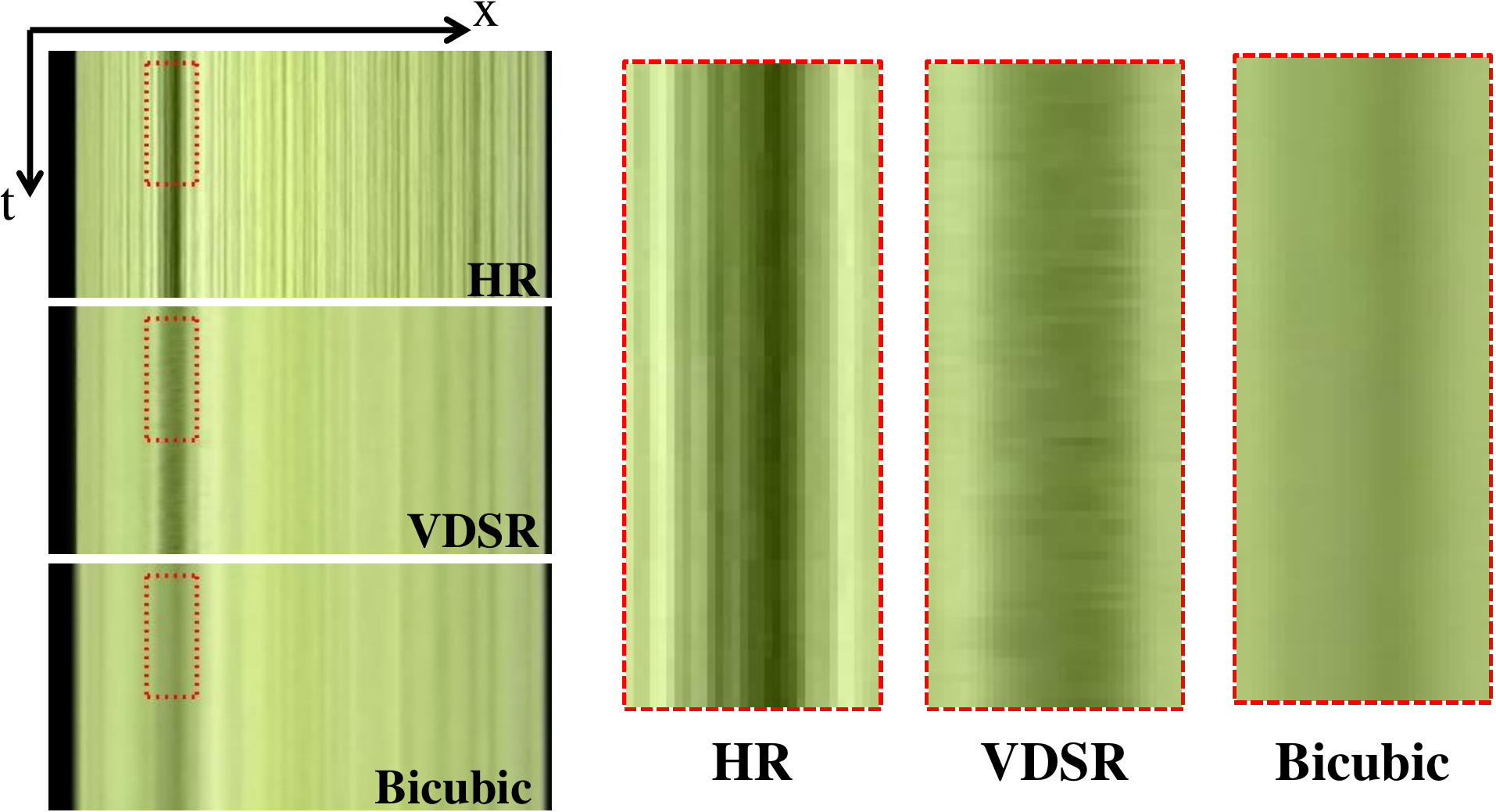}
	\end{minipage}}
	\caption{(a) An example video sequence. We sample one row at the same location (indicated by the red dot line) from each frame and concatenate the rows to produce (b) the temporal profiles. Obviously, bicubic video has the least image details, VDSR video has some details but displays temporal discontinuity that will cause flickering artifacts. In this example, VDSR incurs lower recognition accuracy than bicubic on the temporal stream.}
	\label{fig:artifact}
\end{figure}

The reason for above results should be attributed to VDSR being an image SR network that enhances video frames individually and causes temporal inconsistency.
For high-quality optical flow, we need to ensure the temporal consistency between frames, which has also been studied in previous video SR works.
For example, \cite{caballero2017real,jo2018deep} discussed the temporal consistency and its relation to visible flickering artifacts when displaying SR video.
In Figure \ref{fig:artifact}, we adopt the visualization method known as \emph{temporal profiles} with which \cite{caballero2017real,jo2018deep} display the flickering artifacts.
As seen, VDSR indeed incurs more temporal discontinuity.

\subsubsection{Method}
Through the above observation, we find a relation between optical flow-based recognition accuracy and the temporal consistency in the SR video. Since the existing video SR schemes usually perform SR frame by frame, they have difficulty in guaranteeing the consistency between SR frames. We consider a siamese network reconstructing the consecutive frames together for training video SR.

As our objective is to achieve high quality optical flow, it is straightforward to calculate the optical flow between SR frames and compare it with that between HR frames. However, this would require an optical flow estimation network to support end-to-end training. But recent optical flow networks \cite{hui2018liteflownet,ilg2017flownet,sun2018pwc} are too deep to be efficiently trained with the standard error back-propagation technology. We take a warping approach to estimate the temporal continuity.

\begin{figure}[t]
	\begin{center}
		\includegraphics[width=\linewidth]{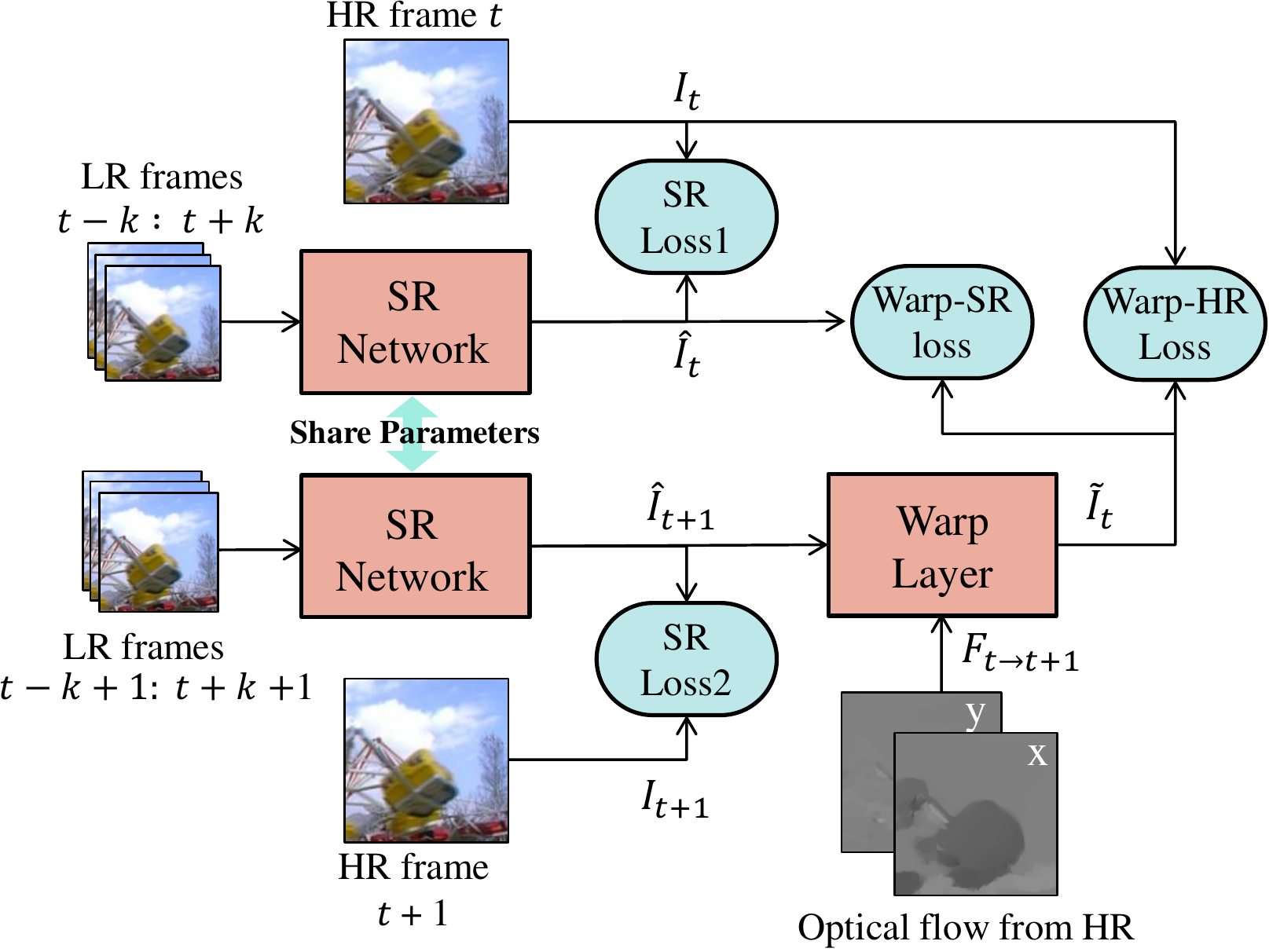}
	\end{center}
	\caption{Our proposed ToSR uses a siamese network for training. We jointly consider two consecutive frames and design four loss terms to ensure the quality of individual frames as well as the temporal consistency between them.}
	\label{fig:ToSR}
\end{figure}

The siamese network for training ToSR is shown in Figure \ref{fig:ToSR}. We use two copies of an SR network to enhance two consecutive frames respectively. First of all, we want to achieve SR frames with high quality, and use two MSE losses for the two frames respectively, \ie $\mathcal{L}_{\textrm{SR}}=\|I_t-\hat{I}_{t}\|_F^2+\|I_{t+1}-\hat{I}_{t+1}\|_F^2$. Moreover, we want to ensure the temporal continuity between SR frames. So we adopt the optical flow from HR video, which can be calculated beforehand, to perform warping between two SR frames.
Let the optical flow be $F_{t\rightarrow t+1}$, we use the relation $\tilde{I}_{t}(p)=\hat{I}_{t+1}(p+F_{t\rightarrow t+1}(p))$ to warp the SR frame $\hat{I}_{t+1}$.
Warping is implemented by bilinear interpolation that is free of parameters.
The warped result $\tilde{I}_{t}$ is compared against both SR and HR frames of the previous timestamp. Accordingly, we define two losses:
$\mathcal{L}_{\textrm{warp-SR}}=\|\hat{I}_{t}-\tilde{I}_{t}\|_F^2$ and 
$\mathcal{L}_{\textrm{warp-HR}}=\|{I}_{t}-\tilde{I}_{t}\|_F^2$.

In summary, the loss function for ToSR is
\begin{equation}
\label{ToSR_loss}
\mathcal{L}_{\textrm{ToSR}}= \alpha \mathcal{L}_{\textrm{SR}}+ \beta \mathcal{L}_{\textrm{warp-SR}}+\gamma \mathcal{L}_{\textrm{warp-HR}},
\end{equation}
where $\alpha$, $\beta$, $\gamma$ are weights.

Any existing image or video SR network can implement ToSR.
We investigate two choices.
The first is based on the VDSR network \cite{kim2016accurate}, which performs SR for frames individually.
The second is based on the VSR-DUF with 16 layers \cite{jo2018deep}, which utilizes multiple LR frames for SR.

The ablation study about the two network structures as well as the proposed loss function is reported in Table \ref{tab:ToSR_Ablation}. The performance of VDSR network is limited by lack of information about adjacent frames while multi-frame SR network performs better on temporal stream. On different networks, our proposed warp loss could benefit to a large extent.

\section{Experiments}
\subsection{Experimental settings}\label{subsec_exp}
\textbf{Datasets.} We perform experiments using three datasets: one natural video dataset CDVL-134 for training SR network, and two datasets, UCF101 and HMDB51, for testing SR with action recognition networks. For the video SR task, there is no commonly used dataset. CDVL-134 is a dataset collected by ourselves from CDVL\footnote{\url{https://www.cdvl.org/}}, and contains 134 natural video sequences with various content, including landscapes, animals, activities and so on. Because the resolution of these sequences varies from 480$\times$360 to 1920$\times$1080, we resize them to around 320$\times$240 (similar to UCF101 and HMDB51) with bicubic interpolation while maintaining their aspect ratios. We further down-sample these resized sequences by a factor of 4 to generate LR video for training. As for UCF101 and HMDB51, they are popular action recognition datasets. The former dataset contains 13,320 video clips belonging to 101 action categories, and the latter is composed of 51 action categories and 6,766 video clips. Both datasets provide three training/testing splits and we here only use the first split as a representative. For more details, please refer to \cite{soomro2012ucf101} and \cite{kuehne2011hmdb} respectively. In our experiment, we use a 4$\times$ down-sampled version of these two datasets as the input LR video for testing, and we use the original resolution of the two datasets (denoted by HR) as a reference.

\begin{table}
	\begin{center}
	\footnotesize\selectfont
		\begin{tabular}{|c|c|c|c|c|}
			\hline  
			Structure &	Warp loss  & Accuracy \\
			\hline
			VDSR & - &  55.1\% \\  
			\hline  
			VDSR & \checkmark &  58.76\% \\ 
			\hline 
			VSR-DUF-16 &  -  & 59.48\% \\  
			\hline 
			VSR-DUF-16 & \checkmark & \textbf{61.5\%} \\  
			\hline  
		\end{tabular}
	\end{center}
	\caption{Ablation study for ToSR using different network structures and different loss functions, with TSN \cite{wang2016temporal} on HMDB51 dataset.}
	\label{tab:ToSR_Ablation}
\end{table}

\textbf{Spatial-oriented SR.} We randomly select HR frames from training video, and use FlowNet2.0 \cite{ilg2017flownet} on them to calculate optical flow which is then converted into weight maps. We crop HR frames and weight maps into 128$\times$128 aligned patches and generate LR patches by bicubic interpolation. In particular, we select 120 frames from each video of CDVL-134 dataset and choose the top 10 crops with the largest area of motion. Manually excluding some obviously low-quality patches, there are totally 144,306 patches for training. We use the deep learning framework PyTorch to perform experiments and the optimization settings, such as learning rate and batch size, are recommended by \cite{wang2018esrgan}. The loss weight used in our SoSR training is $\alpha=1, \beta=1, \gamma=0.005$.

\begin{table*}
	\begin{center}
		\resizebox{2.1\columnwidth}{!}{
			\begin{tabular}{|c||c|c|c|c|c|c||c|c|c|c|c|c|}
				\hline
				&  \multicolumn{6}{c||}{HMDB51} &  \multicolumn{6}{|c|}{UCF101} \\
				\cline{2-13}
				Method &  \multicolumn{3}{c|}{TSN} &  \multicolumn{3}{c||}{ST-Resnet} &  \multicolumn{3}{c|}{TSN} &  \multicolumn{3}{c|}{ST-Resnet} \\
				\cline{2-13}
				& Spatial &  Temporal &  Fusion  & Spatial &  Temporal &  Fusion  & Spatial &  Temporal &  Fusion  & Spatial &  Temporal &  Fusion \\
				\cline{2-13}
				\hline
				Bicubic   & 42.81	  & 56.54	 &63.53    	& 43.59	   & 53.76	  &59.48    & 71.25	    & 81.08    & 87.87   & 72.01	 & 78.28	& 84.62 \\
				\hline
				VDSR \cite{kim2016accurate}  & 46.6	 & 55.1	 & 63.59 & 49.18	& 54.44	& 60.2  & 67.09	& 79.81	& 86.84 & 72.27	& 79.43	& 84.48 \\
				\hline
				RCAN \cite{zhang2018image}	 & 48.76 & 56.8	 & 66.21 & 51.76	& 55.72	& 62.61 & 67.18 & 82.12	& 88    & 72.23	& 80.52	& 85.01 \\
				\hline
				SRGAN \cite{ledig2017photo}	 & 48.82 & 49.87 & 63.01 & 51.41	& 47.22	& 60.85 & 81.33 & 75.45 & 87.55 & 83.31	& 70.16	& 86.97 \\
				\hline
				ESRGAN \cite{wang2018esrgan}	     & 52.48 & 51.5	 & 63.4     & 53.79	& 49.72	& 61.83 & 82.97 & 75.32 & 87.75 & 83.81	& 70.64	& 86.62 \\
				\hline
				SoSR	& \textbf{53.59}	 & 50.26 & 64.51 & \textbf{54.77}	& 48.27 & 63.01 & \textbf{83.11} & 74.1	& 86.63 & \textbf{83.92} & 69.68 & 85.77 \\
				\hline
				\hline
				SPMC \cite{tao2017detail} 	 & 48.95 & 56.41 & 64.31 & 53.14	& 53.53	& 63.66 & 70.42 & 80.19	& 87.15 & 74.45	& 77.44	& 84.09 \\
				\hline
				VSR-DUF-16 \cite{jo2018deep} & 48.37 & 59.48 & 66.08 & 50.62	& 55.07	& 61.11 & 68.56 & 84.89 & 89.36 & 72.11 & 80.06	& 83.9 \\
				\hline
				VSR-DUF-52 \cite{jo2018deep} & 48.5	 & 60.52 & 66.86 & 52.84	& 57.61	& 65.23 & 70.54 & 85.09 & 89.85 & 74.49 & 80.16	& 84.88 \\
				\hline				
				ToSR	& 47.45	& \textbf{61.5}      & 66.08 & 51.54 & \textbf{58.92}   & 64.77	& 64.79 & \textbf{85.29}        & 88.46 & 70.88	& \textbf{81.07} & 83.82 \\
				\hline
				\hline
				SoSR+ToSR      & /    & /	 & \textbf{68.3} &/	    & /	        & \textbf{67.32} & /      & /		&  \textbf{92.13}	&/	    & /	    & \textbf{90.19}  \\
				\hline
				\textit{HR}    & \textit{54.58}	     & \textit{62.16}	        & \textit{69.28}	& \textit{56.01} 	& \textit{59.41}    & \textit{68.1}	 & \textit{86.02}	& \textit{87.63}	& \textit{93.49}    & \textit{88.01}	& \textit{85.71}	& \textit{92.94} \\
				\hline
		\end{tabular}}
	\end{center}
	\caption{Recognition accuracy (\%) of 4$\times$ super-resolved video from UCF101 and HMDB51 dataset using two action recognition network, TSN and ST-Resnet. Number of VSR-DUF \cite{jo2018deep} indicates number of layers. Accuracy of HR video is provided for reference. (Please refer to the supplementary material for PSNR and SSIM results of different methods.)}
	\label{tab:Table1}
\end{table*}

\textbf{Temporal-oriented SR.} All the training samples are prepared similarly as for SoSR, except that TVL$^1$ \cite{zach2007duality} is applied on HR frames to calculate optical flow for warping (for a fair comparison, because the recognition networks use TVL$^1$), and we have 143,250 patch pairs for training, and 10,386 pairs for validation. Our ToSR is implemented on TensorFlow. The initial learning rate is 0.01 and multiplied by 0.1 every 10 epochs as recommended in \cite{jo2018deep}. We use batch size 16 and fine-tune from the model provided by the authors of \cite{jo2018deep}. As for loss weights, we have $\alpha=1, \beta=0.8, \gamma=0.1$.
\subsection{Recognition Results}\label{subsec_res}
All experimental results are obtained by different SR methods with the same scaling factor 4. Baseline methods are four single image SR methods: VDSR \cite{kim2016accurate}, RCAN \cite{zhang2018image}, SRGAN \cite{ledig2017photo}, ESRGAN \cite{wang2018esrgan}, and two video SR methods: SPMC \cite{tao2017detail} and VSR-DUF \cite{jo2018deep}. TSN \cite{wang2016temporal} and ST-Resnet \cite{feichtenhofer2016spatiotemporal} are used to obtain the recognition accuracy, shown in Table \ref{tab:Table1}.

On spatial stream, firstly, comparing VDSR and RCAN, SRGAN and ESRGAN respectively shows that advanced design of network structure could benefit recognition quality of super-resolved video. This result is intuitive because more advanced SR method would generate SR frames with more details and be more helpful to recognition \emph{on average}. Secondly, by comparing VDSR/RCAN and SRGAN/ESRGAN, the former methods optimize MSE only while the latter methods take use of perceptual loss, we can see the perceptual loss could also improve the recognition performance to some extent. Thirdly, our SoSR achieves the highest recognition accuracy even outperforming ESRGAN that is believed to achieve the best perceptual index \cite{wang2018esrgan}; this demonstrates that the SR methods optimized for visual quality are not optimal for action recognition task. For visual analyses please refer to Section \ref{subsec_visual}.

Switching to temporal stream, where it is obvious that SRGAN/ESRGAN perform worse than VDSR/RCAN among single image SR methods. This difference should result from the perceptual loss, as Ledig \etal explained in \cite{ledig2017photo}: MSE-based result is the pixel-wise average of possible results in pixel space, while GAN drives the reconstruction towards the natural image manifold. Accordingly, MSE-based result has better temporal consistency between adjacent SR frames.
Among multi-frame SR methods, VSR-DUF outperforms SPMC significantly, even VSR-DUF-16 beats SPMC with a large gap. This difference may be attributed to the design of network structure. SPMC performs explicit warping with optical flow estimated from LR frames, which may introduce errors and undermine temporal consistency, while VSR-DUF uses 3D convolution directly operating on consecutive LR frames to predict dynamic filters that are then used to up-sample the central LR frame. The structure without explicit motion compensation may be the key for VSR-DUF to achieve good performance. Last, owing to the proposed siamese network, the performance of our ToSR is even better.

Combining the two streams, SoSR plus ToSR gives out the highest accuracy. Results of other combinations are provided in the supplementary material.



\subsection{Analyses}\label{subsec_visual}

\begin{figure}[t]
	\begin{center}
		\includegraphics[width=\linewidth]{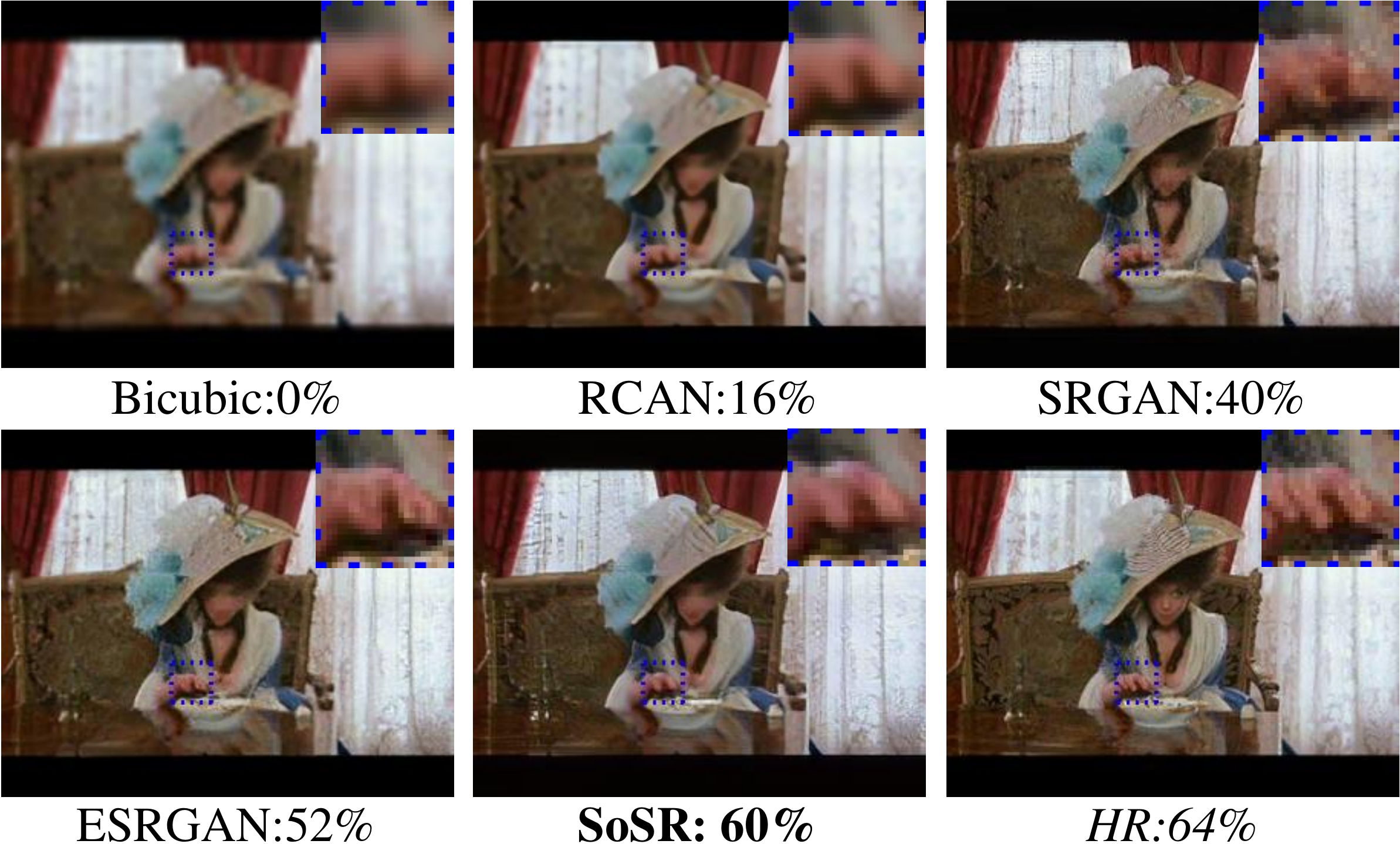}
	\end{center}
	\caption{Visual quality comparison, and the numbers indicate video-level recognition accuracy using TSN. We observe from the SoSR result that the woman's hand looks over-smooth but the background texture appears sharp. This is due to the interaction between WMSE and perceptual losses.}
	\label{fig:v_SoSR}
\end{figure}

Figure \ref{fig:v_SoSR} shows an example for comparing the visual quality as well as recognition accuracy of different SR methods.
By comparing RCAN, SRGAN, ESRGAN results with the HR frame, we can find the recognition accuracy increases as the visual quality improves. When adding SoSR result into consideration, we observe that its visual quality is not consistent at different regions. As shown by the inset, the woman's hand is over-smooth, but the background texture appears quite sharp.
This is due to the interaction between WMSE and perceptual loss: WMSE emphasizes the MSE loss on the regions with large motion, and the MSE loss leads to over-smooth result as claimed in \cite{ledig2017photo}; on the regions with little motion (\eg the background), perceptual loss dominates the optimization target, which produces vivid but not true texture.
In addition, we also observed cases where visual quality and recognition accuracy are not consistent, please refer to the supplementary material.

\begin{figure}[t]
	\begin{center}
		\includegraphics[width=\linewidth]{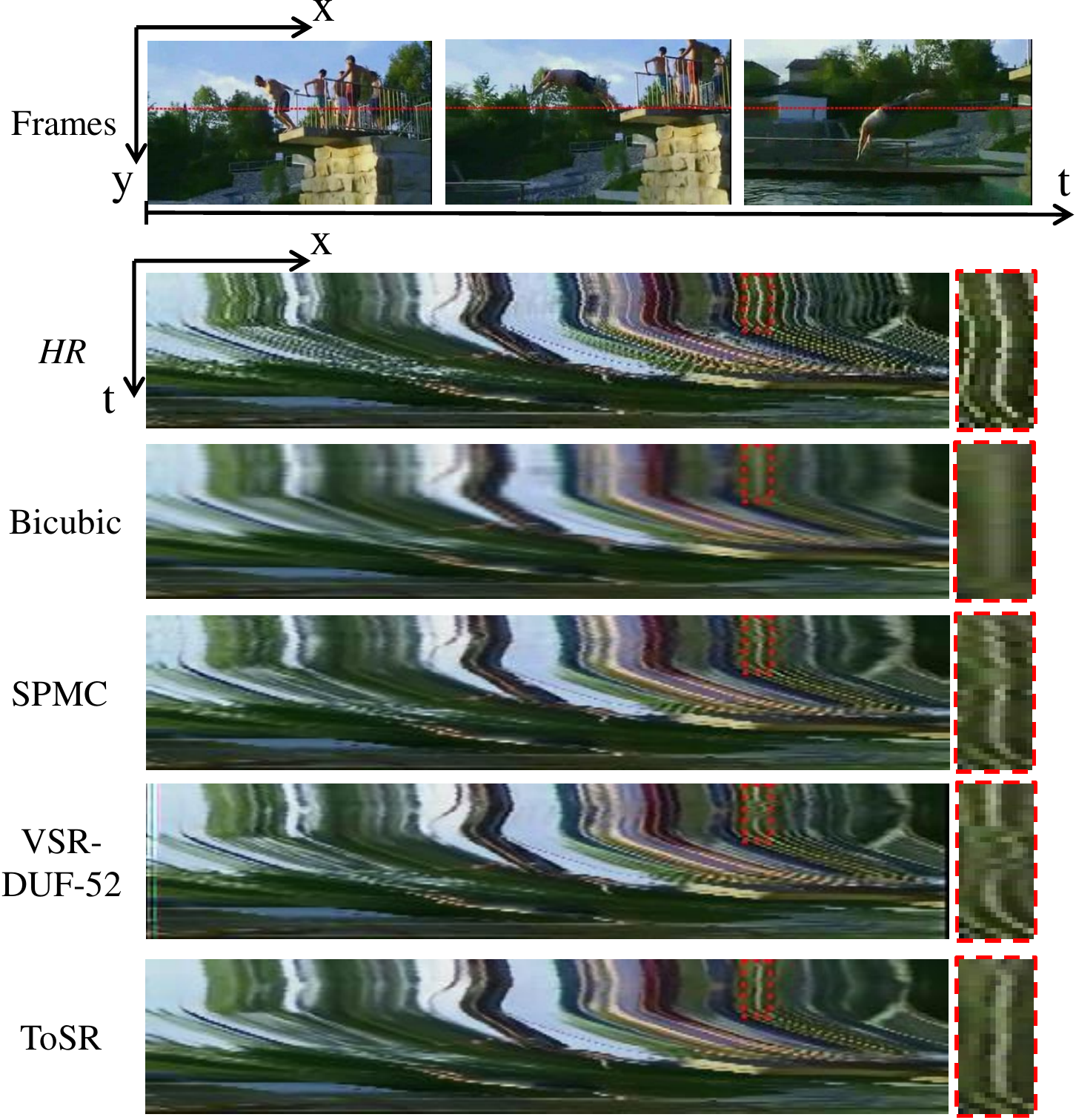}
	\end{center}
	\caption{Temporal profiles of different SR results. Bicubic failed to produce vivid image details while SPMC and VSR-DUF incur obvious temporal discontinuity.}
	\label{fig:v_ToSR2}
	\vspace{-5pt}
\end{figure}

Figure \ref{fig:v_ToSR2} shows temporal profiles of video obtained by different SR methods, from which we can find bicubic interpolated frames do not have enough image details, while SPMC and VSR-DUF results look sharp. However, both of them incur severe temporal discontinuity. Our ToSR produces the best temporal profile. For more visual results about temporal profiles as well as artifacts in optical flow, please refer to the supplementary material.

In all of the previous experiments, we adopt action recognition network pretrained with HR video and evaluate different SR methods.
One may question about this setting and consider whether it would be different if the action recognition network is trained \emph{with video of different resolutions}.
We investigate the joint influence of data augmentation when training recognition network and SR preprocessing when using the trained recognition network.

\begin{figure}[t]
  \begin{center}
    \includegraphics[width=\linewidth]{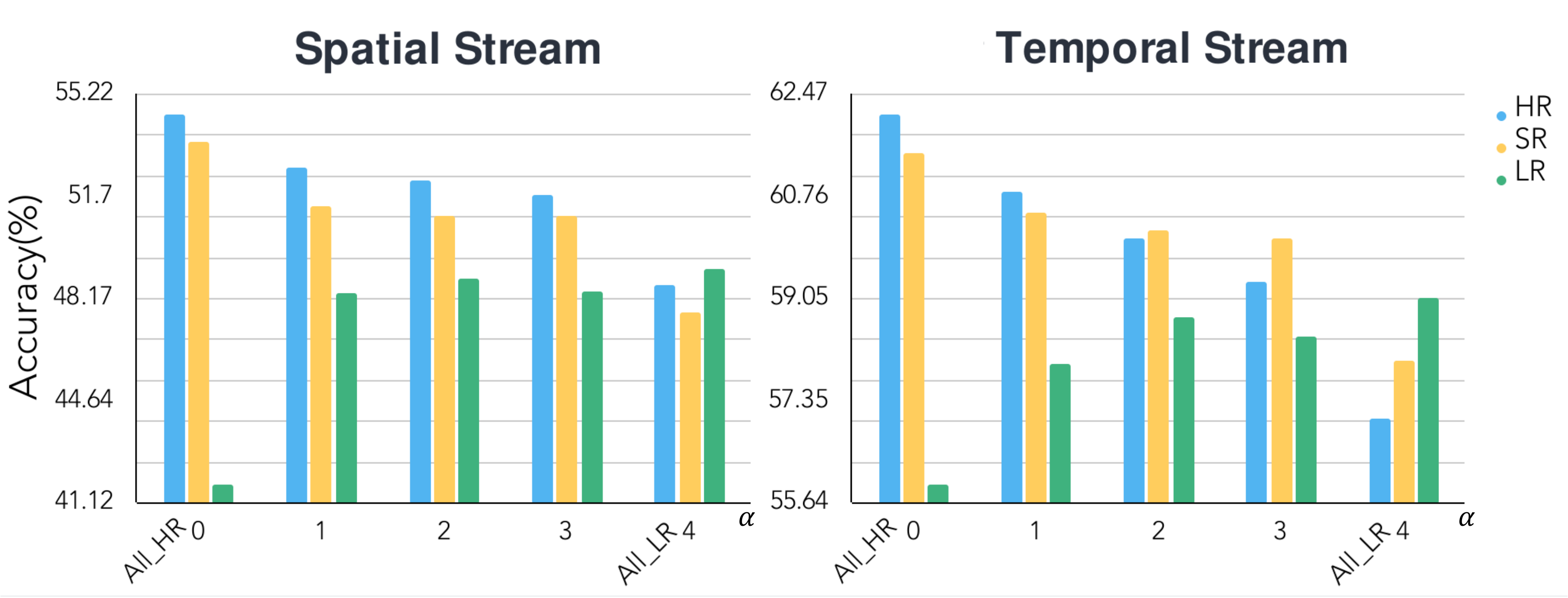}
  \end{center}
  \caption{Recognition accuracy of models trained with different data augmentation configurations (denoted by $\alpha$) and tested on HR, SR, LR video respectively.}
  \label{fig:DataAug}
  \vspace{-5pt}
\end{figure}

In our experiment, we train several TSN models using mixed HR and LR video from the HMDB51 dataset.
Here the LR video is generated by 4$\times$ down-sampling.
The LR and HR video sequences are mixed with a ratio $\alpha : (4-\alpha)$, \eg $\alpha=0$ means HR only and $\alpha=4$ means LR only.
Then we test the recognition performance of each model on HR video, LR video and super-resolved video (using SoSR and ToSR) respectively and report the results in Figure \ref{fig:DataAug}.

Firstly, using data augmentation can make the network pay more attention to the features shared by HR and LR, and improve the performance on LR video. However, the large difference between HR and LR may cause the network to neglect useful features unique to HR, and incur a decline in HR performance. Secondly, the network trained on LR video only ($\alpha=4$) performs the best on LR video input, where the accuracy on LR is even higher than that on HR and SR. But this network performs the worst on HR video input, and should not be a good choice in practice.
Thirdly, excluding the LR only case, there are still cases where SR outperforms HR (on the temporal stream). Thus, we anticipate that a joint consideration of SR network and action recognition network may lead to even better performance, which will be our future work.

Training different models is a straightforward solution for different resolutions, but has several limitations. First, it needs to train and maintain multiple models that can be costly. Second, how to select the appropriate model to match the resolution for a given input video is a problem.
\section{Conclusion}
We consider the video SR problem not for visual quality, but for facilitating action recognition accuracy. Tailored for two-stream action recognition networks, we propose SoSR with optical flow guided weighted MSE loss, and ToSR with a siamese network to emphasize temporal consistency. Experimental results demonstrate the advantages of our proposed SoSR and ToSR methods. In the future, we plan to combine SoSR and ToSR into a single step, and study the tradeoff between visual quality and recognition accuracy.


{\small
\begin{thebibliography}{10}\itemsep=-1pt

\bibitem{blau2018perception}
Yochai Blau and Tomer Michaeli.
\newblock The perception-distortion tradeoff.
\newblock In {\em CVPR}, pages 6228--6237, 2018.

\bibitem{caballero2017real}
Jose Caballero, Christian Ledig, Andrew~P Aitken, Alejandro Acosta, Johannes
  Totz, Zehan Wang, and Wenzhe Shi.
\newblock Real-time video super-resolution with spatio-temporal networks and
  motion compensation.
\newblock In {\em CVPR}, volume~1, pages 4778--4787, 2017.

\bibitem{dai2016image}
Dengxin Dai, Yujian Wang, Yuhua Chen, and Luc Van~Gool.
\newblock Is image super-resolution helpful for other vision tasks?
\newblock In {\em WACV}, pages 1--9, 2016.

\bibitem{diba2017deep}
Ali Diba, Vivek Sharma, and Luc Van~Gool.
\newblock Deep temporal linear encoding networks.
\newblock In {\em CVPR}, pages 2329--2338, 2017.

\bibitem{donahue2015long}
Jeffrey Donahue, Lisa Anne~Hendricks, Sergio Guadarrama, Marcus Rohrbach,
  Subhashini Venugopalan, Kate Saenko, and Trevor Darrell.
\newblock Long-term recurrent convolutional networks for visual recognition and
  description.
\newblock In {\em CVPR}, pages 2625--2634, 2015.

\bibitem{dong2016image}
Chao Dong, Chen~Change Loy, Kaiming He, and Xiaoou Tang.
\newblock Image super-resolution using deep convolutional networks.
\newblock {\em IEEE Transactions on Pattern Analysis and Machine Intelligence},
  38(2):295--307, 2016.

\bibitem{feichtenhofer2016spatiotemporal}
Christoph Feichtenhofer, Axel Pinz, and Richard Wildes.
\newblock Spatiotemporal residual networks for video action recognition.
\newblock In {\em NIPS}, pages 3468--3476, 2016.

\bibitem{feichtenhofer2016convolutional}
Christoph Feichtenhofer, Axel Pinz, and Andrew Zisserman.
\newblock Convolutional two-stream network fusion for video action recognition.
\newblock In {\em CVPR}, pages 1933--1941, 2016.

\bibitem{goodfellow2014generative}
Ian Goodfellow, Jean Pouget-Abadie, Mehdi Mirza, Bing Xu, David Warde-Farley,
  Sherjil Ozair, Aaron Courville, and Yoshua Bengio.
\newblock Generative adversarial nets.
\newblock In {\em NIPS}, pages 2672--2680, 2014.

\bibitem{herath2017going}
Samitha Herath, Mehrtash Harandi, and Fatih Porikli.
\newblock Going deeper into action recognition: A survey.
\newblock {\em Image and Vision Computing}, 60:4--21, 2017.

\bibitem{hui2018liteflownet}
Tak-Wai Hui, Xiaoou Tang, and Chen~Change Loy.
\newblock {LiteFlowNet}: A lightweight convolutional neural network for optical
  flow estimation.
\newblock In {\em CVPR}, pages 8981--8989, 2018.

\bibitem{ilg2017flownet}
Eddy Ilg, Nikolaus Mayer, Tonmoy Saikia, Margret Keuper, Alexey Dosovitskiy,
  and Thomas Brox.
\newblock Flownet 2.0: Evolution of optical flow estimation with deep networks.
\newblock In {\em CVPR}, volume~2, pages 2462--2470, 2017.

\bibitem{ji20133d}
Shuiwang Ji, Wei Xu, Ming Yang, and Kai Yu.
\newblock {3D} convolutional neural networks for human action recognition.
\newblock {\em IEEE Transactions on Pattern Analysis and Machine Intelligence},
  35(1):221--231, 2013.

\bibitem{jo2018deep}
Younghyun Jo, Seoung~Wug Oh, Jaeyeon Kang, and Seon~Joo Kim.
\newblock Deep video super-resolution network using dynamic upsampling filters
  without explicit motion compensation.
\newblock In {\em CVPR}, pages 3224--3232, 2018.

\bibitem{johnson2016perceptual}
Justin Johnson, Alexandre Alahi, and Li Fei-Fei.
\newblock Perceptual losses for real-time style transfer and super-resolution.
\newblock In {\em ECCV}, pages 694--711, 2016.

\bibitem{kappeler2016video}
Armin Kappeler, Seunghwan Yoo, Qiqin Dai, and Aggelos~K Katsaggelos.
\newblock Video super-resolution with convolutional neural networks.
\newblock {\em IEEE Transactions on Computational Imaging}, 2(2):109--122,
  2016.

\bibitem{kim2016accurate}
Jiwon Kim, Jung~Kwon Lee, and Kyoung~Mu Lee.
\newblock Accurate image super-resolution using very deep convolutional
  networks.
\newblock In {\em CVPR}, pages 1646--1654, 2016.

\bibitem{kim2016deeply}
Jiwon Kim, Jung~Kwon Lee, and Kyoung~Mu Lee.
\newblock Deeply-recursive convolutional network for image super-resolution.
\newblock In {\em CVPR}, pages 1637--1645, 2016.

\bibitem{kuehne2011hmdb}
Hildegard Kuehne, Hueihan Jhuang, Est{\'\i}baliz Garrote, Tomaso Poggio, and
  Thomas Serre.
\newblock {HMDB}: A large video database for human motion recognition.
\newblock In {\em ICCV}, pages 2556--2563, 2011.

\bibitem{ledig2017photo}
Christian Ledig, Lucas Theis, Ferenc Husz{\'a}r, Jose Caballero, Andrew
  Cunningham, Alejandro Acosta, Andrew~P Aitken, Alykhan Tejani, Johannes Totz,
  and Zehan Wang.
\newblock Photo-realistic single image super-resolution using a generative
  adversarial network.
\newblock In {\em CVPR}, pages 4681--4690, 2017.

\bibitem{lim2017enhanced}
Bee Lim, Sanghyun Son, Heewon Kim, Seungjun Nah, and Kyoung~Mu Lee.
\newblock Enhanced deep residual networks for single image super-resolution.
\newblock In {\em CVPRW}, number~2, pages 136--144, 2017.

\bibitem{liu2019single}
Ding Liu and Thomas~S Huang.
\newblock Single image super-resolution: From sparse coding to deep learning.
\newblock In {\em Deep Learning through Sparse and Low-Rank Modeling}, pages
  47--86. Elsevier, 2019.

\bibitem{liu2017robust}
Ding Liu, Zhaowen Wang, Yuchen Fan, Xianming Liu, Zhangyang Wang, Shiyu Chang,
  and Thomas Huang.
\newblock Robust video super-resolution with learned temporal dynamics.
\newblock In {\em ICCV}, pages 2507--2515, 2017.

\bibitem{liu2018learning}
Ding Liu, Zhaowen Wang, Yuchen Fan, Xianming Liu, Zhangyang Wang, Shiyu Chang,
  Xinchao Wang, and Thomas~S Huang.
\newblock Learning temporal dynamics for video super-resolution: A deep
  learning approach.
\newblock {\em IEEE Transactions on Image Processing}, 27(7):3432--3445, 2018.

\bibitem{qiu2017learning}
Zhaofan Qiu, Ting Yao, and Tao Mei.
\newblock Learning spatio-temporal representation with pseudo-{3D} residual
  networks.
\newblock In {\em ICCV}, pages 5534--5542, 2017.

\bibitem{shermeyer2018effects}
Jacob Shermeyer and Adam Van~Etten.
\newblock The effects of super-resolution on object detection performance in
  satellite imagery.
\newblock {\em arXiv preprint arXiv:1812.04098}, 2018.

\bibitem{shi2016real}
Wenzhe Shi, Jose Caballero, Ferenc Husz{\'a}r, Johannes Totz, Andrew~P Aitken,
  Rob Bishop, Daniel Rueckert, and Zehan Wang.
\newblock Real-time single image and video super-resolution using an efficient
  sub-pixel convolutional neural network.
\newblock In {\em CVPR}, pages 1874--1883, 2016.

\bibitem{simonyan2014two}
Karen Simonyan and Andrew Zisserman.
\newblock Two-stream convolutional networks for action recognition in videos.
\newblock In {\em NIPS}, pages 568--576, 2014.

\bibitem{soomro2012ucf101}
Khurram Soomro, Amir~Roshan Zamir, and Mubarak Shah.
\newblock {UCF101}: A dataset of 101 human actions classes from videos in the
  wild.
\newblock {\em arXiv preprint arXiv:1212.0402}, 2012.

\bibitem{sun2018pwc}
Deqing Sun, Xiaodong Yang, Ming-Yu Liu, and Jan Kautz.
\newblock {PWC-Net}: {CNNs} for optical flow using pyramid, warping, and cost
  volume.
\newblock In {\em CVPR}, pages 8934--8943, 2018.

\bibitem{sun2015human}
Lin Sun, Kui Jia, Dit-Yan Yeung, and Bertram~E Shi.
\newblock Human action recognition using factorized spatio-temporal
  convolutional networks.
\newblock In {\em ICCV}, pages 4597--4605, 2015.

\bibitem{tai2017image}
Ying Tai, Jian Yang, and Xiaoming Liu.
\newblock Image super-resolution via deep recursive residual network.
\newblock In {\em CVPR}, number~2, pages 3147--3155, 2017.

\bibitem{tai2017memnet}
Ying Tai, Jian Yang, Xiaoming Liu, and Chunyan Xu.
\newblock {MemNet}: A persistent memory network for image restoration.
\newblock In {\em CVPR}, pages 4539--4547, 2017.

\bibitem{tao2017detail}
Xin Tao, Hongyun Gao, Renjie Liao, Jue Wang, and Jiaya Jia.
\newblock Detail-revealing deep video super-resolution.
\newblock In {\em ICCV}, pages 22--29, 2017.

\bibitem{tran2015learning}
Du Tran, Lubomir Bourdev, Rob Fergus, Lorenzo Torresani, and Manohar Paluri.
\newblock Learning spatiotemporal features with {3D} convolutional networks.
\newblock In {\em ICCV}, pages 4489--4497, 2015.

\bibitem{varol2018long}
G{\"u}l Varol, Ivan Laptev, and Cordelia Schmid.
\newblock Long-term temporal convolutions for action recognition.
\newblock {\em IEEE Transactions on Pattern Analysis and Machine Intelligence},
  40(6):1510--1517, 2018.

\bibitem{vidal2018ug}
Rosaura~G Vidal, Sreya Banerjee, Klemen Grm, Vitomir Struc, and Walter~J
  Scheirer.
\newblock {UG2}: A video benchmark for assessing the impact of image
  restoration and enhancement on automatic visual recognition.
\newblock In {\em WACV}, pages 1597--1606, 2018.

\bibitem{wang2016temporal}
Limin Wang, Yuanjun Xiong, Zhe Wang, Yu Qiao, Dahua Lin, Xiaoou Tang, and Luc
  Van~Gool.
\newblock Temporal segment networks: Towards good practices for deep action
  recognition.
\newblock In {\em ECCV}, pages 20--36, 2016.

\bibitem{wang2018two}
Xuanhan Wang, Lianli Gao, Peng Wang, Xiaoshuai Sun, and Xianglong Liu.
\newblock Two-stream {3D} convnet fusion for action recognition in videos with
  arbitrary size and length.
\newblock {\em IEEE Transactions on Multimedia}, 20(3):634--644, 2018.

\bibitem{wang2018esrgan}
Xintao Wang, Ke Yu, Shixiang Wu, Jinjin Gu, Yihao Liu, Chao Dong, Yu Qiao, and
  Chen~Change Loy.
\newblock {ESRGAN}: Enhanced super-resolution generative adversarial networks.
\newblock In {\em ECCVW}, pages 63--79, 2018.

\bibitem{wu2015modeling}
Zuxuan Wu, Xi Wang, Yu-Gang Jiang, Hao Ye, and Xiangyang Xue.
\newblock Modeling spatial-temporal clues in a hybrid deep learning framework
  for video classification.
\newblock In {\em ACM MM}, pages 461--470, 2015.

\bibitem{yue2015beyond}
Joe Yue-Hei~Ng, Matthew Hausknecht, Sudheendra Vijayanarasimhan, Oriol Vinyals,
  Rajat Monga, and George Toderici.
\newblock Beyond short snippets: Deep networks for video classification.
\newblock In {\em CVPR}, pages 4694--4702, 2015.

\bibitem{zach2007duality}
Christopher Zach, Thomas Pock, and Horst Bischof.
\newblock A duality based approach for realtime {TV-L$^1$} optical flow.
\newblock In {\em Joint Pattern Recognition Symposium}, pages 214--223, 2007.

\bibitem{zhang2018image}
Yulun Zhang, Kunpeng Li, Kai Li, Lichen Wang, Bineng Zhong, and Yun Fu.
\newblock Image super-resolution using very deep residual channel attention
  networks.
\newblock In {\em ECCV}, pages 1--16, 2018.

\bibitem{zhou2018mict}
Yizhou Zhou, Xiaoyan Sun, Zheng-Jun Zha, and Wenjun Zeng.
\newblock {MiCT: Mixed 3D/2D} convolutional tube for human action recognition.
\newblock In {\em CVPR}, pages 449--458, 2018.

\end{thebibliography} }
\end{document}